\documentclass{article}
\usepackage{log_2024}						

\usepackage{booktabs}						
\usepackage{multirow}						
\usepackage{amsfonts}						
\usepackage{graphicx}						
\usepackage{duckuments}						

\usepackage{array}
\newcolumntype{H}{>{\setbox0=\hbox\bgroup}c<{\egroup}@{}}

\usepackage[numbers,compress,sort]{natbib}	

\usepackage[textsize=tiny]{todonotes}
\usepackage{makecell} 
\usepackage{pifont}   

\usepackage{cancel}

\title{Revisiting Graph Homophily Measures}

\author[Mikhail Mironov and Liudmila Prokhorenkova]{
Mikhail Mironov\\
Yandex Research\\
\email{mironov.m.k@gmail.com}\And
Liudmila Prokhorenkova\\
Yandex Research\\
\email{ostroumova-la@yandex-team.ru}
}

\usepackage{amsthm}
\theoremstyle{plain}
\newtheorem{theorem}{Theorem}[section]
\newtheorem{proposition}[theorem]{Proposition}

\theoremstyle{definition}
\newtheorem{definition}{Definition}
\newtheorem{property}{Property}

\theoremstyle{remark}

\newcommand{\R}{\mathbb{R}}
\newcommand{\rand}{\mathrm{rand}}
\newcommand{\ep}{\epsilon}

\usepackage{wrapfig}

\begin{document}

\maketitle

\begin{abstract}
Homophily is a graph property describing the tendency of edges to connect similar nodes. There are several measures used for assessing homophily but all are known to have certain drawbacks: in particular, they cannot be reliably used for comparing datasets with varying numbers of classes and class size balance. To show this, previous works on graph homophily suggested several properties desirable for a good homophily measure, also noting that no existing homophily measure has all these properties. Our paper addresses this issue by introducing a new homophily measure~--- \emph{unbiased homophily}~--- that has all the desirable properties and thus can be reliably used across datasets with different label distributions. The proposed measure is suitable for undirected (and possibly weighted) graphs. We show both theoretically and via empirical examples that the existing homophily measures have serious drawbacks while unbiased homophily has a desirable behavior for the considered scenarios. Finally, when it comes to directed graphs, we prove that some desirable properties contradict each other and thus a measure satisfying all of them cannot exist. 
\end{abstract}

\section{Introduction}

Graphs serve as a natural data structure in many areas: social networks, molecules, databases, road traffic, citation networks, etc. Graph nodes often come with labels that characterize them and divide them into several classes. For instance, in a molecular graph, these labels could represent different types of atoms, while in a citation network, they might indicate the scientific disciplines to which each paper belongs. Based on the relationship between graph edges and node labels, graphs are typically categorized as either homophilic or heterophilic~\cite{luan2024heterophilic}. A graph is called \emph{homophilic} if its nodes are more likely to connect with nodes having the same label. For example, a citation network where papers are annotated with their research areas is a homophilic graph since papers tend to cite other papers from the same research area. In contrast, a graph is called \emph{heterophilic} if its nodes are more likely to be connected to nodes with different labels. An example of a heterophilic graph is a supply chain network where suppliers tend to connect to manufacturers but not to other suppliers.

Graph neural networks (GNNs) are the primary modern machine learning tools for working with graphs. The performance of GNNs can vary depending on whether they are applied to homophilic or heterophilic graphs. Earlier studies on GNNs primarily focused on homophilic graphs; however, there is a growing debate regarding the necessity of developing specialized models tailored for heterophilic settings \citep{zhu2020beyond, ma2021homophily, luan2022revisiting, platonov2023critical}. Thus, the notion of homophily and heterophily is important for machine learning on graphs, and it is important to be able to measure how homophilic or heterophilic each graph is.

A \emph{homophily measure} is a graph characteristic that indicates the level of homophily of a given graph. For undirected graphs, many homophily measures were constructed in previous works, e.g., edge homophily, node homophily, class homophily, and adjusted homophily. Since these measures may often disagree on which graph is more homophilic, being able to decide which measure is more reliable is important. In a recent paper, \citet{HomophilyHeterophilyDichotomy} propose the following approach for comparing different homophily measures: first, formulate desirable properties that a reliable measure is expected to satisfy, and then check what measures have these properties. The authors formulate five such desirable properties. In a nutshell, these properties set the range of a homophily measure values, ensure that homophily increases when we add a homophilic or remove a heterophilic edge, and guarantee that the measure is not biased towards a particular number of classes or their size balance. \citet{HomophilyHeterophilyDichotomy} prove that none of the known homophily measures has all five desirable properties and recommend using adjusted homophily since it dominates all other measures. Finally, they pose an open question regarding the existence of a measure that has all five properties.

We solve this open problem by proposing a new homophily measure~--- \emph{unbiased homophily}. It has all five desirable properties and thus is not biased towards a particular number of classes or their size balance. Hence, unbiased homophily can be reliably used across different datasets. To additionally illustrate this, we conduct experiments showing that all known measures have their failure cases, while unbiased homophily works well in all the considered scenarios. The proposed measure can be applied to any undirected graphs, including weighted with non-negative weights.

Then, we extend our research to directed graphs. We note that all the considered desirable properties of a homophily measure can be naturally formulated for the directed case. However, we prove that in this case, the desirable properties contradict each other, thus no homophily measure for directed graphs can have all five of them.  This opens a direction for future research --- we need to rethink the list of desirable properties for the directed case and find a reliable measure.

To sum up, our paper answers the open question of whether there exists a homophily measure satisfying all the desirable properties: positively for undirected and negatively for directed graphs.

\section{Background on Homophily Measures}\label{sec:known-measures}

Our work focuses on \emph{homophily measures} that quantify how well graph structure agrees with node labels. In a homophilic graph, edges tend to connect nodes with the same label. In graph analysis literature, \emph{homophily} is usually referred to as \emph{label assortativity} (we refer to Appendix~\ref{app:assortativity} for a discussion of assortativity measures).

Let us now define homophily measures used in previous studies. We start with notation. Suppose we are given a graph $G = (V,E)$ with nodes $V$, $|V|=n$, and edges $E$. For now, we assume that $G$ is undirected but may include self-loops and multiple edges. For simplicity, we assume that $G$ is unweighted, but all the results and measures easily generalize to the weighted graphs by replacing edge indicators with edge weights. Each node $v$ has a class label $y_v \in \{1,  \dots, m\}$. We denote by $d(v)$ the degree of $v$ and by $N(v)$ the multiset of neighbors of $v$, that is $d(v) = |N(v)|$. We denote by $n_k$ the size of the $k$-th class, i.e., $n_k = |\{v : y_v=k\}|$. Finally, by $D_k$ we denote the total degree of the $k$-th class, i.e., $D_k = \sum \limits_{v:y_v=k} d(v)$.

\textbf{Edge homophily}~\cite{abu2019mixhop,zhu2020beyond} is simply the fraction of homophilic edges: \[h_{edge}=\frac{ |\{\{u,v\} \in E : y_u=y_v\}|} {|E|}.\] 

\textbf{Node homophily}~\cite{pei2020geom} computes how homophilic each node is and then averages the values over all nodes: \[h_{node}=\frac 1 n \sum \limits_{v \in V} \frac{ |\{u \in N(v) : y_u=y_v \}|} {d(v)}.\] 

\textbf{Class homophily}~\citep{lim2021large} sums positive excess homophily in every class: \[ h_{class} = \frac{1}{m-1}  \sum \limits_{k=1}^m \left[ \frac{ \sum \limits_{v:y_v=k} |\{u\in N(v): y_u=y_v\}|} {D_k} - \frac{n_k}{n}\right]_+,\]
where $[x]_+=\mathrm{max}\{x,0\}.$ The factor $\frac{1}{m-1}$ scales $h_{class}$ to the interval $[0,1]$. 

\textbf{Adjusted homophily}~\citep{HomophilyHeterophilyDichotomy} (also known as {\it assortativity coefficient}~\citep{newman2003mixing}) is defined as: 
\begin{equation}\label{eq:h_adj}
h_{adj} = \frac{h_{edge} - \sum \limits_{k=1}^m D_k^2/(2|E|)^2} {1 - \sum \limits_{k=1}^m D_k^2/(2|E|)^2}.
\end{equation}
Here the numerator is the difference between the observed fraction of homophilic edges and its expected value assuming that the edge endpoints are connected randomly. The denominator scales the obtained value so that the maximum achievable homophily equals one. \citet{HomophilyHeterophilyDichotomy} recommend using adjusted homophily instead of other homophily measures since it is the only measure that is unbiased towards particular class size distributions (as it has the constant baseline property, see Section~\ref{sec:homophily-properties} for the discussion).

To outline the scope of the paper, we first note that our work solely focuses on graph-label relations. In other words, we do not consider node features to be a part of a homophily measure. Also, in graph machine learning literature, the term \emph{homophily} is sometimes used with a different meaning~\cite{luan2022revisiting,zheng2024missing}: that a good homophily measure is expected to correlate well with GNN performance. However, \citet{HomophilyHeterophilyDichotomy} suggest separating these concepts since both highly homophilic and highly heterophilic datasets are easy for GNNs and thus other types of measures should be applied for evaluating the simplicity of a dataset for GNNs~\cite{luan2022revisiting,zheng2024missing,luan2024graph}. In this work we also stick to the standard definition of a homophily measure: it evaluates whether similar nodes are connected. In particular, this implies that we do not evaluate and compare homophily measures based on how well they correlate with GNN performance. Instead, we rely on the theoretical approach based on properties that a good homophily measure is expected to satisfy.

\section{Inconsistency of Homophily Measures}
\label{sec:inconsistency}

It turns out that homophily measures discussed in the previous section are often inconsistent with each other. That is, given two graphs $G_1$ and $G_2$, the graph $G_1$ can be more homophilic than $G_2$ according to one homophily measure, while $G_2$ more homophilic than $G_1$ according to another homophily measure. 

\begin{wraptable}[9]{r}{0.45\textwidth}
\centering
\vspace{-16pt}
\caption{Agreement of homophily measures on the AIDS dataset}
\label{tab:homophily_agreement_AIDS}
\begin{tabular}{l|cccc}
\toprule
 & $h_{edge}$ & $h_{node}$ & $h_{class}$ & $h_{adj}$ \\ \midrule
$h_{edge}$  & - & 91\% & 60\% & 73\% \\ 
$h_{node}$  & 91\% & - & 59\% & 74\% \\ 
$h_{class}$ & 60\% & 59\% & - & 71\% \\ 
$h_{adj}$   & 73\% & 74\% & 71\% & - \\ 
\bottomrule
\end{tabular}
\end{wraptable}

To illustrate this inconsistency, we conducted an experiment on the AIDS molecular dataset~\cite{molecular_datasets}. This dataset contains molecular graphs and we consider the types of atoms as node classes. We sampled $1000$ pairs of graphs. For each pair and each homophily measure, either the first graph is more homophilic, the second graph is more homophilic, or they are equally homophilic according to this measure. Thus, for two homophlily measures, we can count the percentage of pairs where the measures agree on what graph is more homophilic. The results are shown in Table~\ref{tab:homophily_agreement_AIDS}. As we can see, while edge and node homophily usually agree with each other, the consistency of all other pairs of measures is below $75\%$, which means that in many cases the measures disagree on which graph is more homophilic.

We conducted similar experiments for the PROTEINS and MUTAG molecular datasets~\cite{molecular_datasets}, synthetic graphs, and real node classification datasets, see Appendix~\ref{sec:inconsistency_tables}. Interestingly, for the MUTAG dataset, the agreement of class homophily with the other measures is below 50\% which means that in most of the cases, the measures are inconsistent. These experiments clearly demonstrate the disagreements between existing homophily measures and thus motivate our study on the comparative analysis of these measures.

\section{Desirable Properties of Homophily Measures}~\label{sec:homophily-properties}
\vspace{-10pt}

Our experiments in the previous section show that known homophily measures are often inconsistent. So how to choose which ones to trust? In other words, can we say that some measures are better at quantifying the level of homophily? To answer this question, we follow previous studies~\cite{HomophilyHeterophilyDichotomy} and formulate what properties a good homophily measure should have. Then, by checking which measures satisfy which properties we can compare these measures. 

In this section, we motivate and discuss such desirable properties. All of these properties (excluding class symmetry) were introduced by \citet{HomophilyHeterophilyDichotomy}. We give an intuitive explanation for each property in this section and refer to Appendix~\ref{sec:formal-homophily-properties} for formal definitions and deeper discussion. To illustrate why each property is needed, we motivate it with a simple example showing desirable and undesirable behavior of the existing homophily measures. Further in the text, we call an edge \emph{homophilic} if it connects nodes with the same label and \emph{heterophilic} otherwise.

\vspace{5pt}
\begin{property}
    \emph{Monotonicity} requires that if we add a homophilic edge or remove a heterophilic edge, homophily of a graph must increase. 
\end{property}

This property is natural to require as it formalizes the meaning of the term `homophily'. 

\begin{wrapfigure}[8]{r}{0.4\textwidth}
  \vspace{-10pt}
  \centering
  \includegraphics[width=0.95\linewidth]{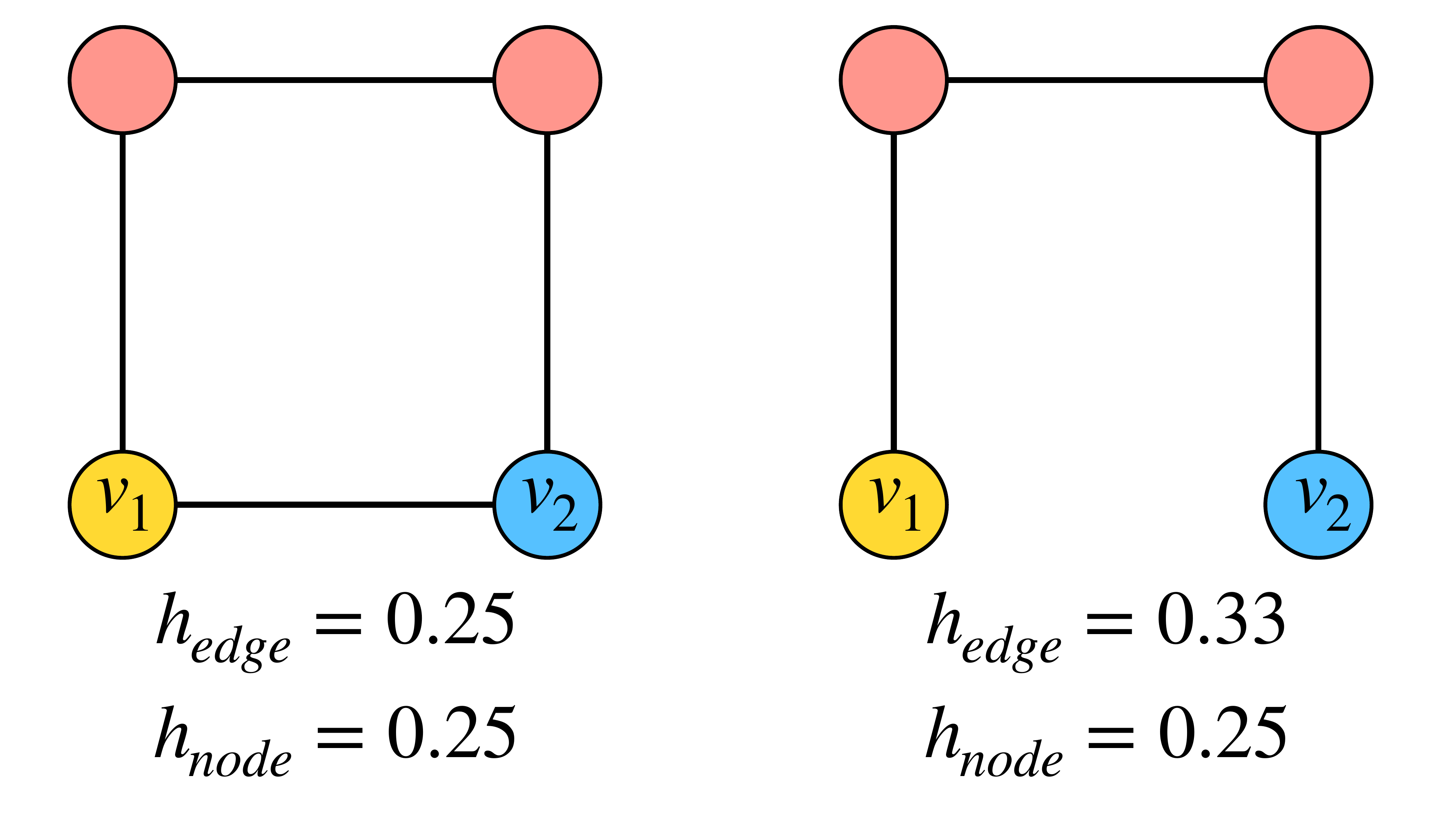} 
\end{wrapfigure}

\paragraph{Example} Consider a graph $G$ and two of its nodes $v_1$ and $v_2$ that are connected and belong to different classes. Assume that $v_1$ and $v_2$ are fully heterophilic, i.e., are not connected to any node of its class. Then, deleting the edge between $v_1$ and $v_2$ does increase edge homophily, but does not change node homophily, see the illustration on the right. We consider the behavior of node homophily to be undesirable in this case.

A more problematic situation is when deleting a heterophilic edge leads to a \emph{decrease} of a homophily measure. We show such an example for adjusted homophily in Appendix~\ref{sec:adj_hom_nonmonotonicity}. 

\vspace{5pt}
\begin{property}
    \emph{Minimal agreement} requires that graphs with only heterophilic edges achieve constant lower bound $R_{min}$ of the homophily measure.
\end{property}

\begin{wrapfigure}[8]{r}{0.4\textwidth}
  \vspace{-10pt}
  \centering
  \includegraphics[width=0.95\linewidth]{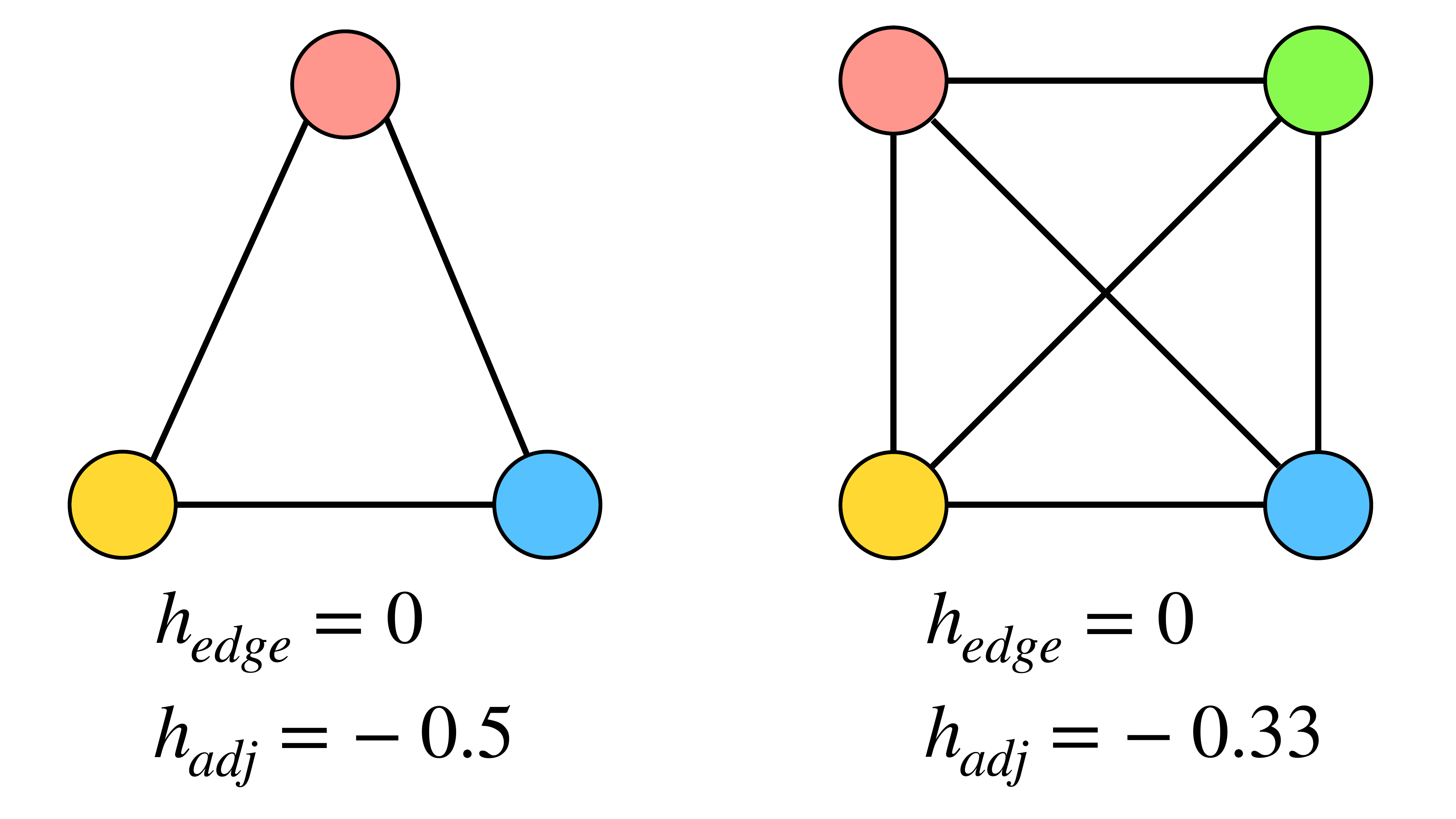} 
\end{wrapfigure}

\paragraph{Example}
Consider graphs $G_1$ with $3$ nodes and $G_2$ with $4$ nodes. Both graphs are complete and the labels of all nodes are different. Clearly, both $G_1$ and $G_2$ are fully heterophilic and we expect a homophily measure to have the lowest possible (and equal) values on these graphs. Edge homophily, node homophily, and class homophily do just that: they take the value $0$ on both these graphs. Yet adjusted homophily takes the value $-0.50$ on $G_1$ and $-0.33$ on $G_2$, which we consider as undesirable behavior.

\vspace{5pt}
\begin{property}
    \emph{Maximal agreement} requires that graphs with only homophilic edges achieve constant upper bound $R_{max}$ of the homophily measure.
\end{property}

Similarly to minimal agreement, this property is natural to require for any good homophily measure. We note that all measures listed in Section~\ref{sec:known-measures} have maximal agreement. 

\vspace{5pt}
\begin{property}
    \emph{Empty class tolerance} requires that if we add a new dummy label that is not present in the graph, the graph homophily does not change.
\end{property}

\paragraph{Example}
Consider a graph and suppose we add a new dummy label that is not present in the graph. That is, the graph does not change, but the formal number of classes $m$ increases by $1$. We expect graph homophily to not change. This property is essential for being able to compare the values of a homophily measure for datasets with different numbers of classes. Clearly, edge homophily, node homophily, and adjusted homophily are not affected by the addition of a dummy label. However, class homophily becomes undefined when we make such an addition since we get $D_k=0$ in the denominator of the corresponding term.

\vspace{5pt}
\begin{property}
    \emph{Constant baseline} requires that if the graph structure is independent of node labels, the homophily measure should be equal to some constant $R_{base}$. We formalize and discuss this property in more detail in Appendix~\ref{sec:formal-homophily-properties}.
\end{property}

\paragraph{Example}
Suppose we have two graphs $G_1$ and $G_2$ with $100$ nodes each. Each graph has two classes, where the class balance of $G_1$ is $50:50$, and the class balance of $G_2$ is $98:2$. Suppose we draw edges according to the Erd{\H{o}}s--R{\'e}nyi model (with self-loops): every two nodes are connected with probability $0.5$ independently of their labels and other edges. Such graphs should be considered neutral with respect to homophily, i.e., neither homophilic nor heterophilic. Thus, we expect both graphs to have the same homophily value (in expectation). The expected value of adjusted homophily is $0$ for both $G_1$ and $G_2$, which we consider as desirable behavior. Yet, the expected value of edge homophily is $0.5$ for $G_1$ and $0.9608$ for $G_2$, which we clearly consider as undesirable behavior.

Finally, let us introduce an additional simple property.

\vspace{5pt}
\begin{property}
    \emph{Class symmetry} requires that if we reorder (or rename) the classes, the value of a homophily measure should not change.
\end{property}
Note that all measures listed in Section~\ref{sec:known-measures} have class symmetry.

The problem raised in previous studies~\cite{HomophilyHeterophilyDichotomy} is that no known measure has all the above properties. Indeed, only adjusted homophily has constant baseline, but it is not monotone for some values of $h_{adj}$, which is a critical drawback. In the next section, we construct a measure that has all the properties.

\section{Homophily Measure Having All Desirable Properties}

In this section, we construct a measure that has all the desirable properties discussed above. Moreover, we require one additional property~--- continuity~--- and we explain why it is necessary below. To formulate this property, we need to first define edge-wise scale-invariant measures.

\subsection{Edge-Wise and Scale-Invariant Measures}\label{-----}

In this section, we say that a \emph{homophily measure} is a function $h$ from the set of all undirected unweighted graphs with labeled nodes to $\R$. A homophily measure is called \emph{edge-wise} if it is a function of the \emph{class adjacency matrix} that we define below.

\vspace{4pt}

\begin{definition}
\label{def:class adjacency matrix} 
For a graph $G$, we define an $m \times m$ \emph{class adjacency matrix} $L_G$ as follows:
\begin{align*}
l_{ii} &= 2 |\{\{u,v\}\in E: y_u= y_v =i\}|\,, \\
l_{ij} &= |\{\{u,v\}\in E: \{y_u, y_v\} = \{i,j\}\}| \text{ for } i \not=j \,.
\end{align*}
\end{definition}
In other words, an edge-wise homophily measure depends on the number of edges between nodes of each pair of classes (or within one class), but it does not take into account how the edges are distributed among individual nodes. Considering the measures listed in Section~\ref{sec:known-measures}, edge homophily and adjusted homophily are edge-wise.

\vspace{4pt}

\begin{definition} 
An edge-wise homophily measure $h$ is called \emph{scale-invariant} if $h(L_G) = h(L_{G'})$ for any graphs $G,G'$ such that $L_G = \beta L_{G'}$ for some $\beta>0$.
\end{definition}

Note that edge homophily and adjusted homophily are scale-invariant.

By definition, edge-wise scale-invariant homophily measures maintain their value under scaling transformations of the class adjacency matrix $L_G$. Let us then normalize $L_G$ as follows.

\vspace{4pt}

\begin{definition}
\label{def:normalized class adjacency matrix} 
For a graph $G$, we define an $m \times m$ \emph{normalized class adjacency matrix} $C_G = \frac{1}{2|E|}L_G:$
\begin{align*}
c_{ii} &= \frac{ |\{\{u,v\}\in E: y_u= y_v =i\}|}{|E|}\,, \\
c_{ij} &= \frac{ |\{\{u,v\}\in E: \{y_u, y_v\} = \{i,j\}\}|}{2|E|} \text{ for } i \not=j\,.
\end{align*}
\end{definition}
For any nonempty graph $G$, its matrix $C_G$ is symmetric, non-negative, and its elements sum to $1$. 

Clearly, every edge-wise scale-invariant homophily measure can be viewed as a function of the normalized class adjacency matrix $C_G$.   
Sometimes (when it is clear from the context), we omit the index $G$ and denote the normalized class adjacency matrix $C_G$ by $C$. Also, we further assume that there are at least two non-zero elements in $C,$ that is, we consider graphs with at least two non-degenerate classes.

For edge-wise scale-invariant homophily measures, we require one more property to hold (in addition to those listed in Section~\ref{sec:homophily-properties}).

\vspace{4pt}

\begin{property}
    \emph{Continuity} requires an edge-wise scale-invariant homophily measure $h$ to be a continuous function of $C$.
\end{property}

This property ensures that small changes in the graph structure (i.e., in the normalized class adjacency matrix) result in small changes in the homophily measure, which is a natural requirement. Note that edge homophily and adjusted homophily are continuous. To show the importance of this property, in Appendix~\ref{sec:measure without continuity} we give an example of a homophily measure that has all desirable properties excluding continuity but demonstrates unwanted behavior, thus proving that continuity must be required.

\paragraph{Formal definition of Properties 1--6.} We formally define all the properties in terms of the normalized class adjacency matrix in Appendix~\ref{sec:formal-homophily-properties}.

\subsection{Unbiased Homophily}\label{sec:example-measure}

In this subsection, we construct a homophily measure that satisfies all the desirable properties. We name this new measure \emph{unbiased homophily}, since by having all the desirable properties it is not biased towards a particular number of classes or their size balance. Hence, unbiased homophily can be reliably used across different datasets: the value of the measure is calibrated for fully homophilic, fully heterophilic, and fully randomized graphs and behaves as desired (due to monotonicity) between these values.\footnote{Note that we do not claim that unbiased homophily is the only possible homophily measure having all the desirable properties. Our goal was to construct at least one measure satisfying these conditions since all the measures from previous studies do not have at least one property.}

We define \emph{unbiased homophily} as:
\begin{equation}\label{eq:unb-homophily}
h_{unb}^{\alpha}(C) := \frac{ \sum \limits_{i <j} (\sqrt{c_{ii}c_{jj}}-c_{ij})} { \sum \limits_{i <j} (\sqrt{c_{ii}c_{jj}}+c_{ij})} +\alpha \min\left( \sum \limits_i \sqrt{c_{ii}},1\right),
\end{equation}
where $\alpha>0$ is any positive constant. Note that the denominator of the first term is always greater than zero, thus $h_{unb}^{\alpha}$ is well-defined. 

\paragraph{Interpretation} First, consider the first term of~\eqref{eq:unb-homophily} since, as discussed below, it is the main ingredient of the measure. Note that if edges were drawn independently of classes, we expect $c_{ij}$ to be equal to $\sqrt{c_{ii}c_{jj}}$ (see the definition of $\rand(C)$ in Appendix~\ref{sec:formal-homophily-properties} for the details). So, $\sqrt{c_{ii}c_{jj}} - c_{ij}$ can be interpreted as the difference between the expected and observed fraction of heterophilic edges between the classes $i$ and $j$ (up to a factor of $2$ since the fraction of such heterophilic edges equals $2c_{ij}$). Summing such differences for all pairs of classes, we interpret the numerator as the difference between the expected and observed fraction of heterophilic edges in the graph (again up to a factor of $2$). The denominator normalizes the obtained value to the interval [-1;1]: when a graph is fully homophilic (i.e., $ c_{ij}=0 \,\, \forall i,j$) the first term equals $\frac{ \sum \limits_{i <j} \sqrt{c_{ii}c_{jj}}} { \sum \limits_{i <j} \sqrt{c_{ii}c_{jj}}}=1$, while for fully heterophilic graphs we get $\frac{ \sum \limits_{i <j} -c_{ij}} { \sum \limits_{i <j} c_{ij}}=-1$.

As we discuss below, the second term $\alpha \min\left( \sum \limits_i \sqrt{c_{ii}},1\right)$ is needed to cover some rare special cases, and in practice, we advise using $\alpha = 0$. So essentially $h_{unb}^{\alpha}$ should be interpreted as the scaled difference between the expected and observed fraction of heterophilic edges.

\paragraph{Theoretical guarantees} The following theorem holds (the proof can be found in Appendix~\ref{sec:proof}).

\vspace{3pt}
\begin{theorem}\label{thm:has-all-properties}
For $\alpha > 0$, the measure $h_{unb}^{\alpha}(C)$ is continuous, has all the desirable properties listed in Section~\ref{sec:homophily-properties}, and its values $R_{max}, R_{base}, R_{min}$ are equal to $1+\alpha, \alpha, -1$, respectively.
\end{theorem}

Note that the desirable properties of unbiased homophily~\eqref{eq:unb-homophily} hold for any $\alpha > 0$. To apply this measure in practice, one has to choose the value of $\alpha$, since different choices may potentially give different results. As we show in the proof of the theorem, the first term of~\eqref{eq:unb-homophily} by itself satisfies all the desirable properties in most cases, and the second term is added to resolve sensitivity issues in some rare special cases. Thus, in practice we recommend taking $\alpha = 0$ as the default value and use:
\begin{equation}\label{eq:unb-homophily-0}
h_{unb}(C) := h_{unb}^{0}(C) = \frac{ \sum \limits_{i <j} (\sqrt{c_{ii}c_{jj}}-c_{ij})} { \sum \limits_{i <j} (\sqrt{c_{ii}c_{jj}}+c_{ij})}.
\end{equation}

\begin{theorem}
The measure $h_{unb}(C)$ is continuous and differentiable and it has all the desirable properties listed in Section~\ref{sec:homophily-properties} except some special cases that only occur when there is at most one non-zero element on the diagonal (i.e., only one class has intra-edges). The values $R_{max}, R_{base}, R_{min}$ are equal to $1, 0, -1$, respectively.
\end{theorem}

The proof of this theorem follows from the proof of Theorem~\ref{thm:has-all-properties}. Note that $h_{unb}$ is differentiable, while $h_{unb}^{\alpha}$ with $\alpha > 0$ is not differentiable at the points where $\sum \limits_i \sqrt{c_{ii}}=1$. 

Thus, unbiased homophily $h_{unb}$ has all the required properties in all cases excluding the case when there is at most one non-zero element on the diagonal (when minimal agreement and monotonicity do not hold). This case rarely appears in real datasets. So, we argue that the simplicity of $h_{unb}$ (in comparison to $h_{unb}^{\alpha}$) outweighs the fact that $h_{unb}$ has slightly weaker theoretical properties. That is why we recommend using unbiased homophily $h_{unb}$ as a homophily measure in practice. See Table~\ref{tab:properties_of_hom_measures} for the list of homophily measures and their properties.

\begin{table}
\centering
\caption{Properties of homophily measures; {\color{green} \checkmark}* means that the property is satisfied always except a rare special case when only one class has intra-edges}
\label{tab:properties_of_hom_measures}
\begin{tabular}{l@{\hskip 5pt} c@{\hskip 5pt} c@{\hskip 5pt} c@{\hskip 5pt} c@{\hskip 5pt} c@{\hskip 5pt} c@{\hskip 5pt} c}
\toprule
 & continuity & \makecell{max \\ agreement} & \makecell{min \\ agreement}  & \makecell{const \\ baseline}  & monotonicity & \makecell{empty class\\ tolerance} & \makecell{class \\ symmetry}  \\ \midrule
$h_{edge}$  & {\color{green} \checkmark} & {\color{green} \checkmark}  & {\color{green} \checkmark}  & {\color{red} \ding{55}}  & {\color{green} \checkmark}  & {\color{green} \checkmark}  & {\color{green} \checkmark} \\ 
$h_{node}$  & n/a & {\color{green} \checkmark}  & {\color{green} \checkmark}  & {\color{red} \ding{55}} & {\color{green} \checkmark}  & {\color{green} \checkmark}  & {\color{green} \checkmark} \\ 
$h_{class}$ & n/a & {\color{green} \checkmark}  & {\color{red} \ding{55}} & {\color{red} \ding{55}}  & {\color{red} \ding{55}}  & {\color{red} \ding{55}}  & {\color{green} \checkmark} \\
$h_{adj}$   & {\color{green} \checkmark} & {\color{green} \checkmark} & {\color{red} \ding{55}}  & {\color{green} \checkmark}  & {\color{red} \ding{55}}  & {\color{green} \checkmark}  & {\color{green} \checkmark}   \\ 
\midrule
$h_{unb}^\alpha$   & {\color{green} \checkmark} & {\color{green} \checkmark}  & {\color{green} \checkmark}  & {\color{green} \checkmark}  & {\color{green} \checkmark}  & {\color{green} \checkmark}  & {\color{green} \checkmark} \\
$h_{unb}$   & {\color{green} \checkmark} & {\color{green} \checkmark}  & {\color{green} \checkmark}*  & {\color{green} \checkmark}  & {\color{green} \checkmark}*  & {\color{green} \checkmark}  & {\color{green} \checkmark} \\
\bottomrule
\end{tabular}
\end{table}

\paragraph{Computing unbiased homophily} The formula~\eqref{eq:unb-homophily-0} is easy to interpret, but there is also an equivalent formula for $h_{unb}$ that is more convenient for computing as it requires only the number of intra-class edges for each class instead of summation over the pairs of classes:
\begin{equation} \label{eq:unb-homophily-1}
  h_{unb}(C) = \frac{ \left(\sum \limits_{i} \sqrt{c_{ii}}\right)^2- 1} { \left(\sum \limits_{i} \sqrt{c_{ii}}\right)^2 + 1 - 2 \sum \limits_{i} c_{ii}}\,.
\end{equation}
We show the equivalence of~\eqref{eq:unb-homophily-0} and~\eqref{eq:unb-homophily-1} in Appendix~\ref{sec:equivalence_formula}.

\paragraph{Unbiased homophily of weighted graphs}

For simplicity of the presentation, all the desirable properties were formulated for unweighted graphs. However, they can be easily rewritten for graphs with positive weights. To do so, in Definition \ref{def:class adjacency matrix} of the class adjacency matrix, instead of the number of edges we use the total weight of the corresponding edges. In Definition \ref{def:normalized class adjacency matrix} of the normalized class adjacency matrix, we replace $|E|$ by the total weight of all edges. Note that the sets of normalized class adjacency matrices in the unweighted and weighted cases are the same (up to the transition from rational to real coefficients by continuity in the unweighted case). Thus, for the weighted case, we can define all properties (see Appendix~\ref{sec:formal-homophily-properties}) in the same way as in the unweighted case. The measures $h_{unb}^{\alpha}$ and $h_{unb}$ for the weighted case have the same properties as for the unweighted case. So, we can use them to measure homophily of weighted graphs. 

\section{Empirical Comparison}

In this section, we empirically compare unbiased homophily with existing homophily measures. First, we consider synthetic examples that demonstrate unwanted behavior of the existing measures, while the results of unbiased homophily are as desired. Then, we compare the level of homophily for different measures on real graph datasets.

\subsection{Synthetic Examples}

In this section, we compare the behavior of homophily measures for simple synthetic datasets.

\paragraph{Example 1.} The first graph $G_1$ is generated by the Erd{\H{o}}s--R{\'e}nyi model with $p=0.5$, it has $2$ classes and the class size balance is $90:10$.

The second graph $G_2$ is generated by the stochastic block model with intra-class edge probability $0.3$ and inter-class probability $0.2$, with $2$ classes and class size balance $50:50$.

\begin{wraptable}[8]{r}{0.5\textwidth}
\centering
\vspace{-15pt}
\captionof{table}{ER with class imbalance vs SBM}
\label{tab:ER_SBM}
\begin{tabular}{l ccc}
\toprule
 & $G_1$, ER & & $G_2$, SBM \\ \midrule
$h_{edge}$  & $0.82$ & \textcolor{red}{>} & $0.6$  \\ 
$h_{node}$  & $0.82$ & \textcolor{red}{>} & $0.6$ \\ 
$h_{class}$ & $0$ & \textcolor{green}{<} &$0.05$\\ 
$h_{adj}$   & $0$ & \textcolor{green}{<} & $0.2$ \\ 
\midrule
$h_{unb}$   & $0$ & \textcolor{green}{<} & $0.38$ \\ 
\bottomrule
\end{tabular}
\end{wraptable}
Clearly, $G_1$ is neutral w.r.t.~homophily since graph edges are independent of node labels. In contrast, $G_2$ exhibits some homophily since intra-class edges have a higher probability. Thus, we expect a good homophily measure to take a higher value on $G_2$. The results for different homophily measures are listed in Table~\ref{tab:ER_SBM}. As we see, edge and node homophily exhibit undesirable behavior for this example.

\paragraph{Example 2.} Both graphs $G_1$ and $G_2$ are complete graphs on $6$ nodes, but $G_1$ has one node in each of $6$ classes, while $G_2$ has two nodes in each of $3$ classes.

\begin{minipage}{0.55\textwidth}
  \centering
  \includegraphics[width=0.75\linewidth]{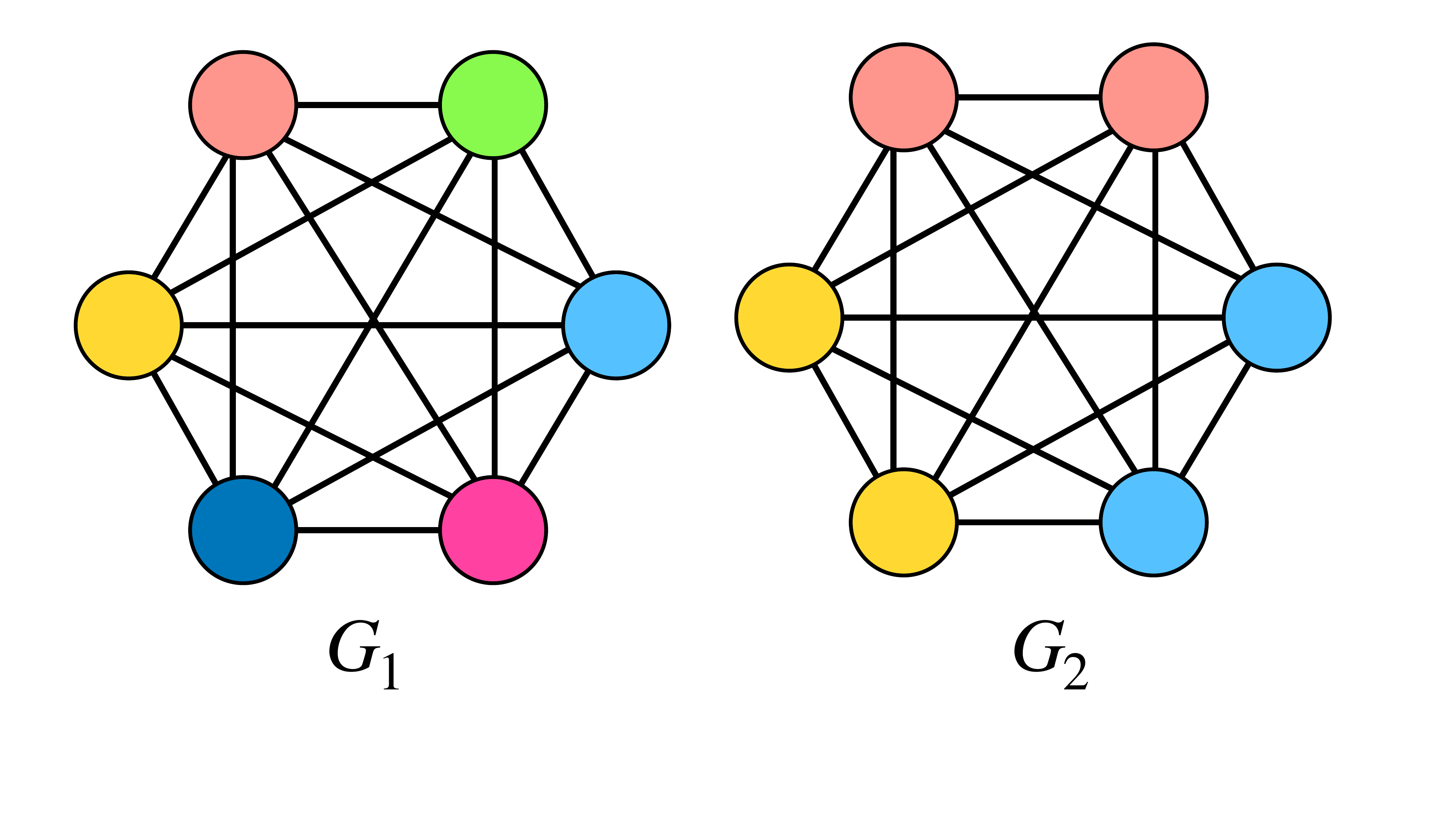} 
  \vspace{-10pt}
  \captionof{figure}{Visualization of $G_1$ and $G_2$ in Example 2}
\end{minipage}
\begin{minipage}{0.45\textwidth}
\centering
\captionof{table}{Homophily values for Example 2}
\label{tab:6complete}
\begin{tabular}{l ccc}
\toprule
 & $G_1$ & & $G_2$ \\ \midrule
$h_{edge}$  & $0$ & \textcolor{green}{<} & $0.2$  \\ 
$h_{node}$  & $0$ & \textcolor{green}{<} & $0.2$ \\ 
$h_{class}$ & $0$ &  \textcolor{red}{=} & $0$\\ 
$h_{adj}$   & $-0.2$ &  \textcolor{red}{=} & $-0.2$ \\ 
\midrule
$h_{unb}$   & $-1$ & \textcolor{green}{<} & $-0.33$ \\ 
\bottomrule
\end{tabular}
\end{minipage}

Clearly, $G_1$ is fully heterophilic, while $G_2$ exhibits some homophily, so we expect a good homophily measure to take a higher value on $G_2$. The results for different homophily measures are listed in Table~\ref{tab:6complete}. As we see, class and adjusted homophily exhibit undesirable behavior.

We see that only unbiased homophily behaves as desired for both of these simple examples. This superiority of unbiased homophily follows from the fact that it satisfies all the desirable properties.

\subsection{Homophily of Real Datasets}

We measure edge, node, class, adjusted, and unbiased homophily for various real datasets that are often used in graph machine learning literature. The description of the datasets can be found in Appendix~\ref{sec:datasets_description}. The results are shown in Table~\ref{tab:real-datasets}.

As expected, node and edge homophily often take higher values, especially on datasets with two classes (that are imbalanced). For instance, for the minesweeper dataset, edge and node homophily are equal to $0.68$, while all the other homophily measures indicate that the homophily level is close to zero (below $0.02$ for all the other measures).

In turn, $h_{class}$, $h_{adj}$, and $h_{unb}$ mostly agree with each other on how homophilic a dataset is. The possible reason for this agreement is that all three measures have been motivated by the constant baseline property. Importantly, $h_{adj}$ and $h_{unb}$ usually agree on whether the homophily value is positive or negative since they both do satisfy constant baseline. 

One can see that the proposed unbiased homophily usually takes higher values than $h_{adj}$ when homophily is positive. For instance, for the amazon-computers dataset, $h_{unb} = 0.91$ and $h_{adj} = 0.68$. Another interesting observation is that for the most heterophilic dataset roman-empire, $h_{adj} = -0.05$ while $h_{unb} = -0.49$, indicating that this dataset is significantly more heterophilic than reported before. This dataset is expected to be quite heterophilic: here the nodes are words, the labels are their syntactic roles, and two nodes are connected if either these words follow each other in the text, or they are connected in the dependency tree of the sentence. This dataset has 18 classes but less than 5\% homophilic edges. Adjusted homophily cannot predict the low score for this dataset since $h_{adj}$ does not satisfy minimal agreement. In contrast, $h_{unb}$ is able to predict a low score.

The difference between edge and unbiased homophily can be illustrated on the pubmed and coauthors-cs datasets. For these datasets, $h_{edge}$ is nearly the same ($0.80$ and $0.81$), but $h_{unb}$ differs significantly: $0.77$ for pubmed and $0.96$ for coauthors-cs. This difference arises because pubmed has $3$ classes, while coauthors-cs has $15$, with relatively balanced class distributions. Unbiased homophily accounts for the fact that in coauthors-cs, having $80\%$ homophilous edges reflects much stronger homophily than in pubmed, due to different class distribution.

\begin{table}
\centering
\caption{Homophily of various datasets}\label{tab:real-datasets}
\begin{small}
\begin{tabular}{l r r c c c c c c c}
\toprule
\textbf{Dataset} & \textbf{$n$} & \textbf{$|E|$} & \textbf{$m$} & $h_{edge}$ & $h_{node}$ & $h_{class}$ & $h_{adj}$ & $h_{unb}$ \\
\toprule
lastfm-asia & 7\,624 & 27\,806 & 18 & 0.8739 & 0.8332 & 0.7656 & 0.8562 & 0.9746 \\
coauthor-cs & 18\,333 & 81\,894 & 15 & 0.8081 & 0.8320 & 0.7547 & 0.7845 & 0.9582 \\
coauthor-physics & 34\,493 & 247\,962 & 5 & 0.9314 & 0.9153 & 0.8474 & 0.8724 & 0.9480 \\
amazon-photo & 7\,650 & 119\,081 & 8 & 0.8272 & 0.8493 & 0.7722 & 0.7850 & 0.9329 \\
cora & 2\,708 & 5\,278 & 7 & 0.8100 & 0.8252 & 0.7657 & 0.7711 & 0.9200 \\
amazon-computers & 13\,752 & 245\,861 & 10 & 0.7772 & 0.8017 & 0.7002 & 0.6823 & 0.9055 \\
facebook & 22\,470 & 170\,823 & 4 & 0.8853 & 0.8834 & 0.8195 & 0.8206 & 0.9033 \\
citeseer & 3\,327 & 4\,552 & 6 & 0.7355 & 0.7166 & 0.6267 & 0.6707 & 0.8520 \\
pubmed & 19\,717 & 44\,324 & 3 & 0.8024 & 0.7924 & 0.6641 & 0.6860 & 0.7653 \\
github & 37\,700 & 289\,003 & 2 & 0.8453 & 0.8011 & 0.3778 & 0.3778 & 0.4969 \\
amazon-ratings & 24\,492 & 93\,050 & 5 & 0.3804 & 0.3757 & 0.1266 & 0.1402 & 0.2769 \\
flickr & 89\,250 & 449\,878 & 7 & 0.3195 & 0.3221 & 0.0698 & 0.0941 & 0.2264 \\
twitch-de & 9\,498 & 153\,138 & 2 & 0.6322 & 0.5958 & 0.1394 & 0.1394 & 0.1566 \\
twitch-pt & 1\,912 & 31\,299 & 2 & 0.5708 & 0.5949 & 0.1196 & 0.1069 & 0.1104 \\
tolokers & 11\,758 & 519\,000 & 2 & 0.5945 & 0.6344 & 0.1801 & 0.0926 & 0.1016 \\
questions & 48\,921 & 153\,540 & 2 & 0.8396 & 0.8980 & 0.0790 & 0.0207 & 0.0583 \\
deezer-europe & 28\,281 & 92\,752 & 2 & 0.5251 & 0.5299 & 0.0304 & 0.0304 & 0.0310 \\
minesweeper & 10\,000 & 39\,402 & 2 & 0.6828 & 0.6829 & 0.0094 & 0.0094 & 0.0145 \\
actor & 7\,600 & 26\,659 & 5 & 0.2167 & 0.2199 & 0.0064 & 0.0028 & 0.0053 \\
genius & 421\,961 & 922\,868 & 2 & 0.5932 & 0.5087 & 0.0229 & -0.0527 & -0.0705 \\
roman-empire & 22\,662 & 32\,927 & 18 & 0.0469 & 0.0460 & 0.0208 & -0.0468 & -0.4913 \\
\bottomrule
\end{tabular}
\end{small}
\end{table}

\section{Homophily for Directed Graphs}

Above, we have only discussed homophily for undirected graphs. The natural next step would be to extend our analysis to directed graphs. The study of edge-wise homophily measures in the directed case is important since these measures have a one-to-one correspondence with classification evaluation measures~\cite{GoodClassificationMeasures}. Thus, if we find a good homophily measure for directed graphs, we also get a good classification evaluation measure that would allow us to compare classification performance for datasets with different numbers of classes. 

In Appendix~\ref{sec:directed}, we reformulate the desirable properties of homophily measures for directed graphs. Unfortunately, it turns out that in the directed case the properties do contradict each other, which we prove in Appendix~\ref{sec:directed}. Thus, no measure satisfying all of them can be constructed. We suggest steps towards modification of these properties to avoid contradictions in the last subsection of Appendix~\ref{sec:directed}.

\section{Conclusion}\label{sec:conclusion}

In this paper, we address the problem of how to choose a reliable homophily measure. We show via a series of examples that previously used measures are flawed and construct a new one~--- unbiased homophily~--- that behaves as desired in all the considered scenarios. We theoretically prove that the proposed measure satisfies the desirable properties introduced in previous studies and some additional ones. Thus, we solve the open problem of whether there exists a homophily measure satisfying all these properties. The proposed measure can be applied to any undirected graphs, and we recommend using it to estimate and compare homophily levels of various graphs in future works. We believe that the proposed measure will become a useful tool for researchers and practitioners. 

In contrast, for directed graphs, we prove that the desirable properties contradict each other, thus no homophily measure can have all the desirable properties. This opens a direction for future research~--- constructing a list of non-contradicting properties and finding the best measure in terms of these properties. Another interesting direction for future studies is to theoretically analyze homophily measures for higher-order relationship networks, such as hypergraphs or simplicial complex networks~\cite{sarker2024higher}.

\bibliographystyle{unsrtnat}

\newpage

\appendix

\section{Graph Assortativity Measures}\label{app:assortativity}

In network science literature, there is a concept related to homophily that is called \emph{assortativity}.  Assortativity is a general concept that is often applied to \emph{node degrees} and the level of assortativity is typically measured by \emph{assortativity coefficient}~\cite{newman2003mixing}. When applied to node labels, the assortativity coefficient reduces to \emph{adjusted homophily}~\citep{platonov2023critical}.

In terms of the class adjacency matrix $C$, the assortativity coefficient (for undirected graphs) can be written as:
\begin{equation}\label{eq:assortativity}
\mathrm{AC} = \frac{\sum \limits_{i} c_{ii} - \sum \limits_{i} \bigg( \sum \limits_{j} c_{ij} \bigg)^2}{1- \sum \limits_{i} \bigg( \sum \limits_{j} c_{ij} \bigg)^2},
\end{equation}
which is equivalent to~\eqref{eq:h_adj}. Thus, the assortativity coefficient (sometimes referred to as \emph{nominal assortativity}) has maximal agreement and constant baseline, but it does not have minimal agreement and monotonicity.

Recently, \citet{karimi2023inadequacy} proposed a new measure which is called {\it adjusted nominal assortativity} that is designed to better handle cases when classes are size imbalanced. The formula of adjusted nominal assortativity can be obtained from~\eqref{eq:assortativity} by dividing each coefficient $c_{ij}$ by $f_if_j$, where $f_i$ is the fraction of nodes in class $i$. That is, the formula of adjusted nominal assortativity for undirected graphs is: 
\[\mathrm{AC}_{adj} = \frac{\sum \limits_{i} \frac{c_{ii}}{f_i f_i} - \sum \limits_{i} \bigg( \sum \limits_{j} \frac{c_{ij}}{f_i f_j} \bigg)^2}{1- \sum \limits_{i} \bigg( \sum \limits_{j} \frac{c_{ij}}{f_i f_j} \bigg)^2} .\]
However, it turns out that scaling the coefficients of nominal assortativity by class sizes prevents the obtained measure from satisfying the desirable properties, as we prove below.

\paragraph{Maximal agreement} For fully homophilic graphs, the value of adjusted nominal assortativity is $\sum \limits_{i} \frac{c_{ii}}{f_i f_i}$. Clearly, by fixing all $c_{ii}$ and changing the number of nodes in classes, we can change the value of $\sum \limits_{i} \frac{c_{ii}}{f_i f_i}$ (given that at least two of $c_{ii}$ are non-zero).
\paragraph{Minimal agreement} Assume that all classes have the same number of nodes, that is, $f_i = \frac{1}{m}$. Then, adjusted nominal assortativity for a fully heterophilic graph becomes:
\[\frac{- \sum \limits_{i} \bigg( \sum \limits_{j} \frac{c_{ij}}{f_i f_j} \bigg)^2}{1- \sum \limits_{i} \bigg( \sum \limits_{j} \frac{c_{ij}}{f_i f_j} \bigg)^2}  = \frac{ - \sum \limits_{i} \bigg( \sum \limits_{j} c_{ij} \bigg)^2 m^4}{1- \sum \limits_{i} \bigg( \sum \limits_{j} c_{ij} \bigg)^2 m^4} .\]
Since for different fully heterophilic graphs the value of $\sum \limits_{i} \bigg(\sum \limits_{j} c_{ij} \bigg)^2$ can be different, the value of adjusted nominal assortativity can also be different.

\paragraph{Constant baseline} Assume that all classes have the same number of nodes, that is, $f_i = \frac{1}{m}$. Then, adjusted nominal assortativity becomes:
\[\frac{\sum \limits_{i} c_{ii} - \sum \limits_{i} \bigg( \sum \limits_{j} c_{ij} \bigg)^2}{1- \sum \limits_{i} \bigg( \sum \limits_{j} c_{ij} \bigg)^2} = \frac{\sum \limits_{i} c_{ii} \, m^2 - \sum \limits_{i} \bigg( \sum \limits_{j} c_{ij} \bigg)^2 m^4}{1- \sum \limits_{i} \bigg( \sum \limits_{j} c_{ij} \bigg)^2 m^4} .\]
Let $C=\rand(C)$ (see the definition of $\rand(C)$ in Appendix~\ref{sec:formal-homophily-properties} for the details), that is, $c_{ii}=\bigg( \sum \limits_{j} c_{ij} \bigg)^2$ for all $i$. Then,
\[\frac{\sum \limits_{i} c_{ii} m^2 - \sum \limits_{i} \bigg( \sum \limits_{j} c_{ij} \bigg)^2 m^4}{1- \sum \limits_{i} \bigg( \sum \limits_{j} c_{ij} \bigg)^2 m^4} = \frac{\sum \limits_{i} c_{ii} m^2 -\sum \limits_{i} c_{ii} m^4}{1- \sum \limits_{i} c_{ii} m^4} = \frac{\sum \limits_{i} c_{ii} (m^2 - m^4)}{1- \sum \limits_{i} c_{ii} m^4}. \]
Since for different $\rand(C)$ the value of $\sum \limits_{i} c_{ii}$ can be different, the value of adjusted nominal assortativity can also be different.

\section{Inconsistency of Homophily Measures}
\label{sec:inconsistency_tables}

In this section, we report the results of additional experiments that are analogous to those in Section~\ref{sec:inconsistency}. As before, we sample $1000$ pairs of graphs and for each pair of homophily measures, we count the percentage of pairs where the measures agree on what graph is more homophilic.

\paragraph{PROTEINS and MUTAG~\cite{molecular_datasets}} The results for these datasets are reported in Tables~\ref{tab:homophily_agreement_PROTEINS} and~\ref{tab:homophily_agreement_MUTAG}. The most notable is the inconsistency between class homophily and other measures in Table~\ref{tab:homophily_agreement_MUTAG}.

\begin{minipage}{0.45\textwidth}
\centering
\captionof{table}{Agreement of homophily measures on the PROTEINS dataset}
\label{tab:homophily_agreement_PROTEINS}
\begin{tabular}{l|cccc}
\toprule
 & $h_{edge}$ & $h_{node}$ & $h_{class}$ & $h_{adj}$ \\ \midrule
$h_{edge}$  & - & 96\% & 76\% & 81\% \\ 
$h_{node}$  & 96\% & - & 76\% & 81\% \\ 
$h_{class}$ & 76\% & 76\% & - & 92\% \\ 
$h_{adj}$   & 81\% & 81\% & 92\% & - \\ 
\bottomrule
\end{tabular}
\end{minipage}
\hspace{0.05\textwidth}
\begin{minipage}{0.45\textwidth}
\centering
\captionof{table}{Agreement of homophily measures on the MUTAG dataset}
\label{tab:homophily_agreement_MUTAG}
\begin{tabular}{l|cccc}
\toprule
 & $h_{edge}$ & $h_{node}$ & $h_{class}$ & $h_{adj}$ \\ \midrule
$h_{edge}$  & - & 97\% & 30\% & 89\% \\ 
$h_{node}$  & 97\% & - & 30\% & 88\% \\ 
$h_{class}$ & 30\% & 30\% & - & 41\% \\ 
$h_{adj}$   & 89\% & 88\% & 41\% & - \\ 
\bottomrule
\end{tabular}
\end{minipage}

\paragraph{Synthetic data} We generate synthetic graphs as follows. We sample the number of classes $m$ uniformly at random from $[2,10]$. Each graph has $100$ nodes, nodes are numbered from $1$ to $100$. We sample $m-1$ different numbers $a_1, \dots, a_{m-1}$ from $[1,99]$ and assign class $i$ to nodes in $(a_{i-1}, a_i]$ (assuming $a_0=1$ and $a_m=100$). For every pair of classes $i \le j$ we sample $p_{ij}$ uniformly at random from $[0,1]$. For every pair of nodes in the graph we draw an edge between them with probability $p_{ij}$ where $i$ and $j$ are classes of these nodes. As a result, we obtain graphs with different community structures and with varying numbers of classes and class size balance. The consistency of the homophily measures on this dataset is shown in Table~\ref{tab:homophily_agreement_synth}.

\begin{minipage}{0.45\textwidth}
\centering
\captionof{table}{Agreement of homophily measures on the synthetic dataset}
\label{tab:homophily_agreement_synth}
\begin{tabular}{l|cccc}
\toprule
 & $h_{edge}$ & $h_{node}$ & $h_{class}$ & $h_{adj}$ \\ \midrule
$h_{edge}$  & - & 97\% & 67\% & 69\% \\ 
$h_{node}$  & 97\% & - & 67\% & 68\% \\ 
$h_{class}$ & 67\% & 67\% & - & 79\% \\ 
$h_{adj}$   & 69\% & 68\% & 79\% & - \\ 
\bottomrule
\end{tabular}
\end{minipage}
\hspace{0.05\textwidth}
\begin{minipage}{0.45\textwidth}
\centering
\captionof{table}{Agreement of homophily measures on the real datasets from Table~\ref{tab:real-datasets}}
\label{tab:homophily_agreement_real_datasets}
\begin{tabular}{l|cccc}
\toprule
 & $h_{edge}$ & $h_{node}$ & $h_{class}$ & $h_{adj}$ \\ \midrule
$h_{edge}$  & -        & 94\%     & 83\%     & 81\%   \\
$h_{node}$  & 94\%     & -        & 85\%     & 82\%   \\
$h_{class}$ & 83\%     & 85\%     & -        & 92\%   \\
$h_{adj}$   & 81\%     & 82\%     & 92\%     & -      \\
\bottomrule
\end{tabular}
\end{minipage}

\paragraph{Node classification datasets} We also conducted a similar measurement on the datasets from Table~\ref{tab:real-datasets}, i.e., we consider all pairs of datasets and check whether two homophily measures agree on which dataset is more homophilic. The results are shown in Table~\ref{tab:homophily_agreement_real_datasets}. Note that the obtained values are (in most cases) higher than in our previous experiments. This difference can be explained by the fact that the datasets in Table~\ref{tab:real-datasets} are very different from each other and, in particular, may have very different homophily levels. When the difference in homophily is significant, it is easy for two measures to agree on which dataset is more homophilic. In contrast, in molecular datasets and synthetic data used for Table~\ref{tab:homophily_agreement_synth}, all graphs are of the same nature and thus are more similar and we expect more inconsistencies. In general, we expect such inconsistency results to highly depend on a dataset. However, note that in all the cases there are many examples of situations when any two measures are inconsistent.

\section{Formal Desirable Properties for Scale-Invariant Homophily Measures}\label{sec:formal-homophily-properties} 

Let us formally define a list of desirable properties a good homophily measure $h$ is expected to satisfy. For convenience of further analysis, we formulate these properties only for edge-wise scale-invariant measures. Also for convenience, we split monotonicity property into two properties: homo-monotonicity and hetero-monotonicity.

Denote by $M_{\mathbb{Q}}$ the set of all matrices which are symmetric, non-negative, with {\it rational} elements that sum to $1$ (and with at least two non-zero elements). Clearly, every element of $M_{\mathbb{Q}}$ corresponds to the normalized class adjacency matrix of some graph, and all matrices corresponding to graphs belong to $M_{\mathbb{Q}}$. Therefore, a homophily measure $h$ can be considered as a function on $M_{\mathbb{Q}}$.

Let us now define the continuity property.

\paragraph{Continuity} A homophily measure $h$ is a continuous function.

This property ensures that small changes in the graph structure (i.e., in the normalized class adjacency matrix) result in small changes in the homophily measure, which is a natural requirement. Note that all edge-wise scale-invariant measures listed in Section~\ref{sec:known-measures} are continuous. To show the importance of this property, in Appendix~\ref{sec:measure without continuity} we give an example of a homophily measure that has all desirable properties excluding continuity but demonstrates unwanted behavior, thus proving that continuity must be required.

Denote by $M_{\mathbb{R}}$ the set of all symmetric non-negative matrices with {\it real} elements that sum to $1$ (and with at least two non-zero elements). Since $M_{\mathbb{Q}}$ is a dense subset of $M_{\mathbb{R}}$, any continuous function $h$ on $M_{\mathbb{Q}}$ can be uniquely extended to a continuous function on $M_{\mathbb{R}}$. Hence, if a homophily measure $h$ is continuous, it can be uniquely extended to a continuous function on $M_{\mathbb{R}}$. Thus, we further assume that continuous homophily measures are defined on $M_{\mathbb{R}}$, i.e., $C$ can be any element of~$M_{\mathbb{R}}$.

\paragraph {Maximal agreement} There exists $R_{max} \in \R$ such that $h(C) \le R_{max}$ and equality holds iff $\sum \limits_{i=1}^m c_{ii} = 1$. In other words, if all edges of a graph are homophilic, then $h(C)=R_{max}$, and if a graph has at least one heterophilic edge, then $h(C)<R_{max}$.

\paragraph {Minimal agreement} There exists $R_{min} \in \R$ such that $h(C) \ge R_{min}$ and equality holds iff $\sum \limits_{i=1}^m c_{ii} = 0$. In other words, if all edges of a graph are heterophilic, then $h(C)=R_{min}$, and if a graph has at least one homophilic edge, then $h(C)>R_{min}$.

\paragraph {Constant baseline} This property ensures that a homophily measure is not biased towards graphs with small or large number of classes or particular class size distributions. To achieve this, the property requires that if a graph structure is independent of labels, then homophily should be equal to some constant~$R_{base}$. 

To formalize this property, for every $C_G$, we construct a new normalized class adjacency matrix $\rand(C_G)$, corresponding to a graph whose structure is independent of labels, while the class degrees distribution is the same as for $C_G$. After that, we require $\forall \, C_G: h(\rand(C_G)) = R_{base}$. 

\begin{definition}
For a normalized class adjacency matrix $C_G$, let $a_i := \sum \limits_{k=1}^m c_{ik} = \frac{D_i}{2|E|}$. Then, we define $\rand(C_G)$ as: $\rand(C_G)_{ij}:=a_i a_j$.
\end{definition}

Let us motivate this definition. Suppose we fix the degrees and labels of all nodes in $G$ and redraw edges between the nodes uniformly at random. That is, every half-edge has an equal probability to form an edge with any other half-edge. It is easy to see that the expected fraction of homophilic edges in class $i$ is $\frac{D_i}{2|E|} \cdot \frac{D_i}{2|E|} = a_i^2$ (up to a negligible term that is usually ignored). Similarly, the expected fraction of heterophilic edges between classes $i$ and $j$ is $2\cdot \frac{D_i}{2|E|} \cdot \frac{D_j}{2|E|} = 2 a_i a_j$. Thus, the elements of $\rand(C_G)$ correspond to the expected fractions of edges of each type under the procedure when we randomly redraw all edges in a graph $G$. Note that column-wise and row-wise sums of $\rand(C_G)$ remain the same as in $C_G$.

We see that the structure of $\rand(C_G)$ is independent of labels. Thus, it is natural to say that $\rand(C_G)$ is neither homophilic nor heterophilic for any $C_G$. This motivates the following definition.

\vspace{4pt}

\begin{definition}[Constant baseline]
A homophily measure $h$ has constant baseline if there exists $R_{base} \in \R$ such that for any $C_G$ we have $h(\rand(C_G)) = R_{base}$. 
\end{definition}

\paragraph{Homo-monotonicity} This property requires that adding homophilic edges increases the value of the measure. Let us reformulate this requirement in terms of the normalized class adjacency matrix $C$. If we add a homophilic edge within a class $i$, then in Definition~\ref{def:normalized class adjacency matrix}, the denominator $|E|$ gets replaced by $|E|+1$, the numerator of $c_{ii}$ increases by $1$, and the numerators of all other elements stay the same. Thus, when we add a homophilic edge within class $i$, all elements of $C$ are multiplied by $\frac{|E|}{|E|+1} = 1 - \frac{1}{|E|+1}$, and after that $c_{ii}$ is increased by $\frac{1}{|E|+1}$. Due to scale invariance, we can make $|E|$ arbitrarily large and thus make $\frac{1}{|E|+1}$ arbitrarily close to zero. Using the fact that we can add several homophilic edges one by one and the continuity property, we can essentially replace $\frac{1}{|E|+1}$ by any number from $(0,1)$, from which the following definition follows.

\begin{definition}[Homo-monotonicity] Suppose $C$ is not fully homophilic, i.e., $\sum \limits_{i=1}^m c_{ii} < 1$. Then, for any $i$ and any $0<\ep<1$ we have $h((1-\ep)C +\ep E_{ii}) > h(C)$, where $E_{ii}$ is a matrix unit (it has $1$ at the position $(i,i)$ and zeros everywhere else). 
\end{definition}

In other words, if not all edges of $G$ are homophilic and we add one or several homophilic edges to $G$, then the resulting graph $G’$ satisfies $h(G’) > h(G)$.

\paragraph {Hetero-monotonicity} This property requires that removing heterophilic edges increases the value of the measure. Similarly to the above, when we delete a heterophilic edge between classes $i$ and $j$, all elements of $C$ are multiplied by $\frac{|E|}{|E|-1} = 1 + \frac{1}{|E|-1}$, and after that $c_{ij}$ and $c_{ji}$ are decreased by $\frac{1}{2(|E|-1)}$.

\begin{definition}[Hetero-monotonicity] Suppose $C$ is not fully heterophilic, i.e., $\sum \limits_{i=1}^m c_{ii} > 0$. Then, for any $i \not= j$ and any $\epsilon$ such that $0<\ep \le 2(1+\epsilon) c_{ij}$ we have 
\[h\bigg((1+\ep)C - \frac{\ep}{2} E_{ij} - \frac{\ep}{2} E_{ji}\bigg) > h(C),
\] 
where $E_{ij}$ is a matrix unit (it has $1$ at the position $i,j$ and zeros everywhere else). 
\end{definition}

In other words, if not all edges of $G$ are heterophilic and we remove one or several heterophilic edges from $G$, then the resulting graph $G’$ satisfies $h(G’) > h(G)$. 

\paragraph {Empty class tolerance} If we pad a matrix $C$ with a row and a column of zeros, the homophily of the matrix does not change. In other words, if we add to $G$ a new dummy label that is not present in the graph, then the resulting graph $G’$ satisfies $h(G') = h(G)$.
     
\paragraph{Class symmetry} Given any $m \times m$ matrix and any permutation $\sigma \in \mathbb{S}_m$, we can simultaneously permute the rows and columns by $\sigma.$ A homophily measure $h$ is called class-symmetric iff for any $C$ and any $\sigma$ we have $h(\sigma(C)) = h(C).$ In other words, the measure $h$ is invariant to permuting (renaming) the classes. Note that all existing measures have class-symmetry.

\section{Homophily Measure Without Continuity Property}\label{sec:measure without continuity}

In this section, we give an example of a homophily measure that has all the desirable properties excluding continuity. We show that it exhibits unwanted behavior and this justifies that continuity is indeed a necessary property.

Consider the following homophily measure:
\begin{align*}
    h(C) &= \sum \limits_i \sqrt{c_{ii}} - 1 \text{ if }  \sum \limits_i \sqrt{c_{ii}} \le 1,\\
    h(C) &= \sum \limits_i c_{ii} \text{ if } \sum \limits_i \sqrt{c_{ii}} > 1.
\end{align*}

Here is an informal idea of the example. The term $\sum \limits_i \sqrt{c_{ii}}$ has minimal agreement and constant baseline, but it does not have maximal agreement. Also, it has hetero-monotonicity everywhere, but its homo-monotonicity is sometimes violated when $\sum \limits_i \sqrt{c_{ii}} > 1$. So, we cut it at $\sum \limits_i \sqrt{c_{ii}} = 1$ to keep all desirable properties and avoid problems with maximal agreement and homo-monotonicity. After that, for $\sum \limits_i \sqrt{c_{ii}} > 1$ we use $\sum \limits_i c_{ii}$ (i.e., edge homophily), which has all desirable properties but does not have constant baseline. The problem with constant baseline is avoided since all $\rand(C)$ graphs (which constant baseline cares about) satisfy $\sum \limits_i \sqrt{c_{ii}} = 1$.

Thus, the measure $h(C)$ is artificially constructed from two measures one of which is known to have no constant baseline. As a result, when a graph is homophilic ($h(C)>0$), $h(C)$ is equal to edge homophily which is known to be not a good measure. This is achieved since we have a discontinuity at $h(C) = 0$.

Indeed, let us consider the following normalized class adjacency matrices:
\vspace{-15pt}
\begin{center}
\[
\renewcommand{\arraystretch}{1.5} 
\begin{array}{cc}
L_1=
\begin{bmatrix}
\frac{1}{4}  & \frac{1}{4} \\
\frac{1}{4}  & \frac{1}{4} \\
\end{bmatrix},
&
\renewcommand{\arraystretch}{1.5} 
L_2=
\begin{bmatrix}
\frac{1}{4} +\ep & \frac{1}{4} -\ep \\
\frac{1}{4} - \ep & \frac{1}{4} +\ep \\
\end{bmatrix} \,,
\end{array}
\]
\end{center}
where $\ep>0$ is a small positive number. 

Since for $L_1$ we have  $\sum \limits_i \sqrt{c_{ii}} = \frac{1}{2}+\frac{1}{2} = 1$, we have $h(L_1) = \sum \limits_i \sqrt{c_{ii}}-1=0$. Since for $L_2$ we have $\sum \limits_i \sqrt{c_{ii}} = \sqrt{\frac{1}{4}+\ep}+\sqrt{\frac{1}{4}+\ep} > 1$, we have $h(L_2) = \sum \limits_i c_{ii} = \frac{1}{2}+2\ep$. Thus, $h(L_1)$ and $h(L_2)$ differ by at least $\frac{1}{2}$, which is an unwanted behavior, since $L_1$ and $L_2$ themself are very close to each other. In particular, for $\ep \to 0$,  the matrix $L_2$ tends to $\rand(L_2)$, but $h(L_2)$ does not tend to $R_{base} = 0$, which indicates that the measure is biased.

Now let us briefly discuss why $h$ indeed has all the desirable properties except continuity.

\paragraph {Continuity}  Not satisfied at the points where $\sum \limits_i \sqrt{c_{ii}}=1$.

\paragraph{Minimal and maximal agreement} The measure $h$ has minimal and maximal agreement with $R_{min}=-1$ and $R_{max}=1$.

\paragraph{Homo- and hetero-monotonicity} The term $\sum \limits_i \sqrt{c_{ii}}$ has homo- and hetero-monotonicity when $\sum \limits_i \sqrt{c_{ii}} \le 1$, as we prove in Appendix~\ref{sec:proof} (since this term is proportional to the second term of unbiased homophily). The term $\sum \limits_i c_{ii}$ has homo- and hetero-monotonicity.

\paragraph {Constant baseline} Suppose $C=\rand(C)$. Then, $\sum \limits_i \sqrt{c_{ii}}= \sum \limits_i \sqrt{a_i^2} = \sum \limits_i a_i =1$. Therefore, $h(C)=\sum \limits_i \sqrt{c_{ii}}=1$. Thus, constant baseline is satisfied with $R_{base} = 0.$

\paragraph {Empty class tolerance} Satisfied, since both $\sum \limits_i \sqrt{c_{ii}}$ and $1+ \sum \limits_i c_{ii}$ do not change after padding a matrix $C$ with a row and a column of zeros. 
     
\paragraph {Class-symmetry}  Satisfied, since both $\sum \limits_i \sqrt{c_{ii}}$ and $1+ \sum \limits_i c_{ii}$ are invariant to permuting (renaming) the classes.

\section{Proof of Theorem~\ref{thm:has-all-properties}}\label{sec:proof}

In this section, we prove that the measure 
\[h_{unb}^{\alpha}(C) := \frac{ \sum \limits_{i <j} (\sqrt{c_{ii}c_{jj}}-c_{ij})} { \sum \limits_{i <j} (\sqrt{c_{ii}c_{jj}}+c_{ij})} +\alpha \min\left( \sum \limits_i \sqrt{c_{ii}},1\right), \,\, \alpha>0
\]
has all the desirable properties listed in Appendix~\ref{sec:formal-homophily-properties}, and its values $R_{max}, R_{base}, R_{min}$ are equal to $1+\alpha, \alpha, -1$, respectively.

\paragraph {Maximal agreement} Suppose $\sum \limits_{i=1}^m c_{ii} = 1$ (and therefore all $c_{ij}=0$). Then the first term is equal to $1$. Since $0\le c_{ii} \le 1$, we have $\sqrt{c_{ii}} \ge c_{ii}$, thus $\sum \limits_i \sqrt{c_{ii}} \ge \sum \limits_i c_{ii} =1$ and $\min( \sum \limits_i \sqrt{c_{ii}},1)=1$. Therefore, the second term is equal to $\alpha$, and homophily is equal to $1+\alpha$.

If $\sum \limits_{i=1}^m c_{ii} < 1$ (and therefore at least one $c_{ij}>0$), then the first term is less than $1$ and the second term is less than or equal to $\alpha$. Thus, maximal agreement is satisfied with $R_{max} = 1+\alpha.$

\paragraph {Minimal agreement}  If $\sum \limits_{i=1}^m c_{ii} = 0$ (and therefore at least one $c_{ij}>0$), then the first term is equal to $-1$ and the second term is equal to $0$. If $\sum \limits_{i=1}^m c_{ii} > 0$, then the first term is greater than or equal to $-1$ and the second term is greater than $0$. Thus, minimal agreement is satisfied with $R_{min} = -1$.

\paragraph {Constant baseline} Suppose $C=\rand(C)$. Then, the first term is equal to $0$ since the numerator of the first term is equal to 
\[\sum \limits_{i <j} \big(\sqrt{c_{ii}c_{jj}}-c_{ij}\big) = \sum \limits_{i <j} \bigg(\sqrt{a_i^2 a_j^2} - a_i a_j\bigg)=\sum \limits_{i <j} (0) = 0.\]
The second term is equal to $\alpha$ since \[\sum \limits_i \sqrt{c_{ii}}= \sum \limits_i \sqrt{a_i^2} = \sum \limits_i a_i =1.\]
Thus, constant baseline is satisfied with $R_{base} = 0+\alpha = \alpha.$

\paragraph {Homo-monotonicity} W.l.o.g.~we can assume that we increase $c_{11}$. That is, all elements of $C$ are multiplied by $(1-\ep)$ and after that the element $(1-\ep)c_{11}$ is increased by $\ep$. We need to prove that after that the homophily measure has increased, given that $C$ has at least one non-zero non-diagonal element.

Let us see what happens to the first term. Multiplying all elements by $(1-\ep)$ does not change the first term, since both numerator and denominator are multiplied by $(1-\ep)$. The subsequent addition of $\ep$ to $(1-\ep)c_{11}$ increases both numerator and denominator by $\delta$, where $\delta =0$ if $c_{22}= \dots=c_{nn}=0$, and $\delta>0$ if at least one of $c_{22}, \dots, c_{nn}$ is not zero. 

Let us prove that if $\delta>0$, then the first term has increased. Denote the numerator by $N$, the denominator by $D$:
\[\frac{N+\delta}{D+\delta} > \frac{N}{D} \ \  \Leftrightarrow \ \ (N+\delta) D > N (D+\delta) \ \ 
    \Leftrightarrow \ \  D\delta > N \delta \ \ \Leftrightarrow \ \  D>N,\]
    
where we used the facts that $D$, $D+ \delta$, and $\delta$ are positive. The last inequality $D>N$ holds for 
all $C$ with at least one non-zero non-diagonal element. Thus, we proved that if $\delta>0$, then the first term has increased. Clearly, if $\delta = 0$, the first term does not change.

Let us see what happens to the second term. Instead of adding $\ep$, we calculate the derivative w.r.t.~$\ep$ at $\ep = 0$:
\begin{multline*}
\frac{\partial }{\partial \ep} \left( \sqrt{(1-\ep) c_{11} +\ep} + \sqrt{(1-\ep) c_{22}} + \ldots + \sqrt{(1-\ep) c_{nn}} \right) \\ = \frac {1}{2} \left( \frac{1-c_{11}}{\sqrt{c_{11}}} - \frac{c_{22}}{\sqrt{c_{22}}} - \ldots - \frac{c_{nn}}{\sqrt{c_{nn}}}\right) = \frac {1}{2} \left(\frac{1}{\sqrt{c_{11}}} - \sum \limits_i \sqrt{c_{ii}} \right)\,.
\end{multline*}
As long as $1 \ge \sum \limits_i \sqrt{c_{ii}}$, this derivative is positive, since $\frac{1}{\sqrt{c_{11}}} > 1 \ge \sum \limits_i \sqrt{c_{ii}}$. Thus, the second term has a positive derivative with respect to $\epsilon$ when $\sum \limits_i \sqrt{c_{ii}}<1$ and zero derivative when $\sum \limits_i \sqrt{c_{ii}} > 1$.

Now let us combine our results for the first and second terms to prove that homo-monotonicity holds. If $\delta > 0$, then the first term is increasing and the second term is non-decreasing, thus homo-monotonicity holds. If $\delta = 0,$ then the only possibly non-zero diagonal element is $c_{11}$ and $c_{11}<1$, thus $\sum \limits_i \sqrt{c_{ii}} = \sqrt{c_{11}}<1$. Therefore, the first term stays constant and the second term increases, thus homo-monotonicity holds. This concludes the proof that $h_{unb}^{\alpha}$ satisfies the homo-monotonicity property.

\paragraph {Hetero-monotonicity} W.l.o.g.~we can assume that $c_{12}$ decreases. That is, all the elements of $C$ are multiplied by $(1+\ep)$ and after that the element $(1+\ep)c_{12}$ is decreased by $\ep$. We need to prove that after that the homophily measure has increased, given that $C$ has at least one non-zero diagonal element.

Let us see what happens to the first term. Multiplying all the elements by $(1+\ep)$ does not change the first term since both the numerator and denominator are multiplied by $(1+\ep)$. The subsequent subtraction of $\ep$ from $(1+\ep)c_{12}$ increases the numerator and decreases the denominator by $\delta =\ep>0$. Denote the numerator by $N$ and the denominator by $D$:
\[\frac{N+\delta}{D-\delta} > \frac{N}{D} \ \  \Leftrightarrow \ \ (N+\delta) D > N (D-\delta) \ \ 
    \Leftrightarrow \ \  D\delta > - N \delta \ \ \Leftrightarrow \ \  D > -N,\]
where we use the facts that $D, D- \delta$, and $\delta$ are greater than $0.$ The last inequality $D> - N$ holds for 
all $C$ with at least two non-zero diagonal elements. Thus, we proved that if $C$ has at least two non-zero diagonal elements, then the first term has increased. Clearly, if $C$ has exactly one non-zero diagonal element, then the first term stays equal to $-1$ and thus does not change.

Let us see what happens to the second term. The expression $\sum \limits_i \sqrt{c_{ii}}$ will be multiplied by $\sqrt{1+\ep}$. Therefore, the second term will increase if $\sum \limits_i \sqrt{c_{ii}}<1$ or stay constant if $\sum \limits_i \sqrt{c_{ii}} \ge 1$. Clearly, if $C$ has exactly one non-zero diagonal element, then $\sum \limits_i \sqrt{c_{ii}}<1$, thus the second term increases. 

Now let us combine our results for the first and second terms to prove that hetero-monotonicity holds. If $C$ has exactly one non-zero diagonal element, then the first term stays constant and the second term increases. If $C$ has at least two non-zero diagonal elements, then the first term increases and the second term non-decreases. This concludes the proof that $h_{unb}^{\alpha}$ satisfies the hetero-monotonicity property.

\paragraph {Empty class tolerance} Satisfied, since both first and second terms do not change after padding a matrix $C$ with a row and a column of zeros. 
     
\paragraph {Class-symmetry}  Satisfied, since both first and second terms are invariant to permuting (renaming) the classes.

\paragraph {Continuity}  Satisfied, since both first and second terms are continuous.

\section{Ajusted Homophily Non-Monotonicity Example}
\label{sec:adj_hom_nonmonotonicity}

Adjusted homophily does not have minimal agreement and it does not have monotonicity when the value of adjusted homophily is low. Let us give an example of unwanted behavior corresponding to the lack of monotonicity.

Consider the following class adjacency matrix. We delete the heterophilic edge between classes $1$ and $2$ and see how adjusted homophily changes:
\begin{center}
\[
\renewcommand{\arraystretch}{1.5} 
\begin{bmatrix}
0   & \cancel{1}  & 0 & 0  \\
\cancel{1}   &  0 & 0 & 0   \\
0  & 0  & 2 & 8  \\
0 & 0 & 8 & 2 \\
\end{bmatrix}.
\]
\end{center}
Before deleting the edge the adjusted homophily was equal to $\frac{ \frac{2}{11} - \frac{1+1+100+100}{22^2}} { 1 - \frac{1+1+100+100}{22^2}} \approx -0.404$. After deleting the edge the adjusted homophily is equal to $\frac{ \frac{2}{10} - \frac{100+100}{20^2}} { 1 - \frac{100+100}{20^2}} = -0.6$. Thus, the adjusted homophily has decreased which contradicts hetero-monotonicity. In contract, the unbiased homophily $h_{unb}$ has increased from $\frac{2-8-1}{2+8+1} \approx -0.636$ to $\frac{2-8}{2+8} = -0.6$. For $h_{unb}^\alpha$ we have an increase from $-0.636 + 0.603\alpha$ to $-0.6+0.632\alpha$.

\section{Equivalence of Two Formulas for $h_{unb}$}
\label{sec:equivalence_formula}

Here is the derivation that proves that the formulas~\eqref{eq:unb-homophily-0} and~\eqref{eq:unb-homophily-1} for $h_{unb}$ are equivalent:
\begin{multline}
    h_{unb}(C) = \frac{ \sum \limits_{i <j} (\sqrt{c_{ii}c_{jj}}-c_{ij})} { \sum \limits_{i <j} (\sqrt{c_{ii}c_{jj}}+c_{ij})} = 
\frac{ 2\sum \limits_{i <j} (\sqrt{c_{ii}c_{jj}}-c_{ij})} { 2\sum \limits_{i <j} (\sqrt{c_{ii}c_{jj}}+c_{ij})} = 
\frac{ \sum \limits_{i \not= j} (\sqrt{c_{ii}c_{jj}}-c_{ij})} { \sum \limits_{i \not= j} (\sqrt{c_{ii}c_{jj}}+c_{ij})}= \\
= \frac{ \left(\sum \limits_{i} \sqrt{c_{ii}}\right)^2- \sum \limits_{i} c_{ii} - \sum \limits_{i \not= j} c_{ij}} { \left(\sum \limits_{i} \sqrt{c_{ii}}\right)^2- \sum \limits_{i} c_{ii} + \sum \limits_{i \not= j} c_{ij}} = \frac{ \left(\sum \limits_{i} \sqrt{c_{ii}}\right)^2- 1} { \left(\sum \limits_{i} \sqrt{c_{ii}}\right)^2 + 1 - 2 \sum \limits_{i} c_{ii}}.
\end{multline}

\section{Comparing Unbiased and Adjusted Homophily}

In this section, we additionally illustrate the difference between unbiased and adjusted homophily with a simple synthetic example.

Here, for a fixed number of classes $m$ and a chosen value of $h_{unb}$, we construct a normalized class adjacency matrix (assuming that homophilic/heterophilic edges are evenly distributed between classes/class pairs), and calculate $h_{adj}$ based on it. Thus, we can see how adjusted homophily changes if we fix unbiased homophily but change the number of classes.

\begin{table}[htbp] 
\centering \caption{Adjusted homophily for different values of $h_{\text{unb}}$ and $m$}
\resizebox{\textwidth}{!}{
\begin{tabular}{c c c c c c c c c c c c} 
\toprule $m, h_{unb}$ & $-1.0$ & $-0.8$ & $-0.6$ & $-0.4$ & $-0.2$ & $0.0$ & $0.2$ & $0.4$ & $0.6$ & $0.8$ & $1.0$ \\ \midrule 2 & -1.000 & -0.800 & -0.600 & -0.400 & -0.200 & 0.000 & 0.200 & 0.400 & 0.600 & 0.800 & 1.000 \\
3 & -0.500 & -0.421 & -0.333 & -0.235 & -0.125 & 0.000 & 0.143 & 0.308 & 0.500 & 0.727 & 1.000 \\
4 & -0.333 & -0.286 & -0.231 & -0.167 & -0.091 & 0.000 & 0.111 & 0.250 & 0.429 & 0.667 & 1.000 \\
5 & -0.250 & -0.216 & -0.176 & -0.129 & -0.071 & 0.000 & 0.091 & 0.211 & 0.375 & 0.615 & 1.000 \\
6 & -0.200 & -0.174 & -0.143 & -0.105 & -0.059 & 0.000 & 0.077 & 0.182 & 0.333 & 0.571 & 1.000 \\
7 & -0.167 & -0.145 & -0.120 & -0.089 & -0.050 & 0.000 & 0.067 & 0.160 & 0.300 & 0.533 & 1.000 \\
8 & -0.143 & -0.125 & -0.103 & -0.077 & -0.043 & 0.000 & 0.059 & 0.143 & 0.273 & 0.500 & 1.000 \\
9 & -0.125 & -0.110 & -0.091 & -0.068 & -0.038 & 0.000 & 0.053 & 0.129 & 0.250 & 0.471 & 1.000 \\
10 & -0.111 & -0.098 & -0.081 & -0.061 & -0.034 & 0.000 & 0.048 & 0.118 & 0.231 & 0.444 & 1.000 \\
\bottomrule \end{tabular}}
\end{table}

We make the following observations. First, when $h_{unb} = 1$, we always have $h_{adj} = 1$. This is expected since this case corresponds to graphs having only homophilic edges and both measures have maximal agreement. Second, when $h_{unb} = 0$, we always have $h_{adj} = 0$. This is also expected since both measures have constant baseline.

Then, when $h_{unb} = -1$, one can see that $h_{adj}$ increases with the number of classes. Since $h_{unb} = -1$ corresponds to graphs containing only heterophilic edges, this shows the lack of minimal agreement for $h_{adj}$.

We also see that when $m=2$, the measures are equal to each other (in our synthetic example). Then, for heterophilic graphs ($h_{unb} < 0$) and fixed $h_{unb}$, $h_{adj}$ increases with the number of classes, while for homophilic graphs ($h_{unb} > 0$) and fixed $h_{unb}$, $h_{adj}$ decreases with the number of classes.

\section{Homophily Measures for Directed Graphs}\label{sec:directed} 

In this section, we reformulate the desirable properties of a homophily measure for directed graphs. After that, we prove that these properties contradict each other. Thus, no measure satisfying all of them can be constructed. In Subsection~\ref{subsec:discussion_directed}, we suggest steps towards modification of these properties to avoid contradictions. 

\subsection{Desirable Properties of Homophily Measures}

In this section, a homophily measure is a function $h$ from the set of all directed unweighted graphs (that may include self-loops and multiple edges) with labeled nodes to $\R$.

\vspace{3pt}

\begin{definition}
For a directed graph $G$, we define an $m \times m$ \emph{class adjacency} matrix $L_G$ by $l_{ij} = |\{(u,v)\in E: (y_u, y_v) = (i,j)\}|$.
\end{definition}

The \emph{normalized class adjacency matrix} $C_G$ is then defined as $\frac{1}{|E|} L_G$. This matrix can be non-symmetric and it has non-negative elements summing to one. As before, we consider only $C_G$ with at least two non-zero elements.

Below we consider only \emph{edge-wise} and \emph{scale-invariant} homophily measures that are defined exactly as in the undirected case. We define the properties \emph{maximal agreement}, \emph{minimal agreement}, \emph{homo-monotonicity}, \emph{empty class tolerance}, \emph{class-symmetry}, and \emph{continuity} as in the undirected case (see Appendix~\ref{sec:formal-homophily-properties}). Let us now define \emph{hetero-monotonicity} and \emph{constant baseline}.

\textbf{Hetero-monotonicity \,\,} As in the undirected case, if not all edges of $G$ are heterophilic and we remove one or several heterophilic edges from $G$, then the resulting graph $G’$ should satisfy $h(G’) > h(G)$.  Formally, let $C$ be not fully heterophilic, i.e., $\sum \limits_{i=1}^m c_{ii} > 0$. Then, for any $i \not= j$ and any $\epsilon$ such that $0<\ep \le (1+\epsilon) c_{ij}$, we have $h((1+\ep)C - {\ep}E_{ij}) > h(C)$.

\textbf{Constant baseline \,\,} As in the undirected case, for every $C_G$, we define the normalized class adjacency matrix $\rand(C_G)$ corresponding to a graph whose structure is independent of labels, while class degrees distribution is the same as for $C_G$. After that, we require $\forall \, C_G: h(\rand(C_G)) = R_{base}$. 

\begin{definition}
For a normalized class adjacency matrix $C_G$, let $a_i := \sum \limits_{k=1}^m c_{ik}$ and $b_j := \sum \limits_{k=1}^m c_{kj}$. Then, we define $\rand(C_G)$ as: $\rand(C)_{ij}:=a_i b_j$.
\end{definition}

To motivate this definition, suppose we fix the in-degrees, out-degrees, and labels of all nodes in $G$ and redraw edges between the nodes uniformly at random. That is, every out-half-edge has an equal probability of forming an edge with any in-half-edge. Similarly to the undirected case, we observe that the expected fraction of homophilic edges in class $i$ is then $a_i b_i$ and the expected fraction of heterophilic edges from class $i$ to class $j$ is $a_i b_j$. Thus, the elements of $\rand(C_G)$ correspond to the expected fractions of edges of each type under the procedure when we randomly redraw all edges in the graph $G$. Note that column-wise and row-wise sums of $\rand(C_G)$ remain the same as in $C_G$.

\vspace{4pt}

\begin{definition}[Constant baseline]
A homophily measure $h$ has constant baseline if there exists $R_{base} \in \R$ such that for any $C_G$ we have $h(\rand(C_G)) = R_{base}$. 
\end{definition}

\subsection{Contradiction}

It turns out that this set of properties is internally contradictory. The following two propositions hold.

\vspace{4pt}

\begin{proposition} \label{prop:const vs min}
    The constant baseline and minimal agreement properties contradict each other.
\end{proposition}
\paragraph{Proof.} Suppose we have a graph $K$ where some classes have only out-edges and all the other classes have only in-edges. Suppose we also have $K=\rand(K)$, thus constant baseline requires the homophily of this graph to be equal to $R_{base}$. On the other hand, minimal agreement requires the homophily of this graph to be equal to $R_{min}$ since all edges of this graph are heterophilic. This easily leads to a contradiction, as we see in the example below: here constant baseline requires $h(K)=h(L)=R_{base}$ and minimal agreement requires $h(K)=R_{min}<h(L)$, which gives a contradiction.
\vspace{-20pt}
\begin{center}
\[
\renewcommand{\arraystretch}{1.5} 
\begin{array}{cc}
K = \begin{bmatrix}
0 & 0 & \frac{1}{4}  & \frac{1}{4} \\
0 & 0 & \frac{1}{4}  & \frac{1}{4}  \\
0 & 0 & 0 & 0 \\
0 & 0 & 0 & 0 \\
\end{bmatrix},
&
\renewcommand{\arraystretch}{1.5} 
L = \begin{bmatrix}
\frac{1}{4}  & \frac{1}{4} & 0 & 0 \\
\frac{1}{4}  & \frac{1}{4} & 0 & 0  \\
0 & 0 & 0 & 0 \\
0 & 0 & 0 & 0 \\
\end{bmatrix}.
\end{array}
\]
\end{center}

\vspace{4pt}

\begin{proposition} \label{prop:const vs hetero-mono}
    The constant baseline and hetero-monotonicity properties contradict each other.
\end{proposition} 
\paragraph{Proof.} Suppose we have graphs $T=\rand(T)$ and $L=\rand(L)$ such that $L$ can be obtained from $T$ by deleting several heterophilic edges. Then, constant baseline requires $h(T)=h(L)=R_{base}$ and hetero-monotonicity requires $h(T)<h(L)$, which gives a contradiction. One such example is shown below, where we replace all non-zero elements in the fourth column and third row of $T$ with zeros, obtaining matrix $L$:
\vspace{-20pt}
\begin{center}
\[
\renewcommand{\arraystretch}{1.5} 
\begin{array}{cc}
T = \begin{bmatrix}
\frac{1}{9}   & \frac{1}{9}  & 0 & \cancel{\frac{1}{9}}  \\
\frac{1}{9}   & \frac{1}{9}  & 0 & \cancel{\frac{1}{9}}   \\
\cancel{\frac{1}{9}}  & \cancel{\frac{1}{9}}  & 0 & \cancel{\frac{1}{9}}  \\
0 & 0 & 0 & 0 \\
\end{bmatrix},
&
\renewcommand{\arraystretch}{1.5} 
L = \begin{bmatrix}
\frac{1}{4}  & \frac{1}{4} & 0 & 0 \\
\frac{1}{4}  & \frac{1}{4} & 0 & 0  \\
0 & 0 & 0 & 0 \\
0 & 0 & 0 & 0 \\
\end{bmatrix}.
\end{array}
\]
\end{center}

\paragraph{Not necessarily edge-wise scalar-invariant measures} We gave the examples above to prove the contradiction between the desirable properties, while the properties were formulated for edge-wise scalar-invariant measures. Let us give a sketch of the proof that the same examples work for any measure that is not necessarily scalar-invariant or even edge-wise.

Indeed, consider our example which shows the contradiction between constant baseline and minimal agreement. Consider two graphs satisfying constant baseline, with the corresponding normalized class adjacency matrices $K$ and $L$. Since all edges of the first graph are heterophilic and not all edges of the second graph are heterophilic, the minimal agreement property requires that homophily of the first graph is $R_{min}$ and homophily of the second is greater than $R_{min}$. Yet, constant baseline requires the homophily of both of them to be equal to $R_{base}$, thus giving a contradiction.

Consider our example that shows a contradiction between constant baseline and hetero-monotonicity. Consider any graph satisfying constant baseline, corresponding to the normalized class adjacency matrix $T$. Delete in this graph all edges with at least one end in classes $3$ and $4$, getting the second graph with the normalized class adjacency matrix $L$. Constant baseline requires the homophily of both of these graphs to be equal to $R_{base}$. Yet we have deleted only heterophilic edges, thus hetero-monotonicity requires the homophily of the second graph to be greater than homophily of the first graph, thus giving a contradiction.

\subsection{Discussion}
\label{subsec:discussion_directed}

As shown above, the list of desirable properties for the directed case is self-contradicting. Hence, it is natural to try to modify these properties to avoid contradictions.

The contradiction between constant baseline and minimal agreement properties is quite severe. This contradiction occurs since there can be cases when $C$ is fully heterophilic and $C=\rand(C)$. A possible way to resolve this problem is to simply allow $h$ to be undefined (and discontinuous) for all such $C$.

To resolve the contradiction between constant baseline and hetero-monotonicity, we can rethink the property of monotonicity. Note that our current definition of constant baseline is naturally formulated in terms of normalized class adjacency matrices, while homo- and hetero-monotonicity are formulated in terms of edges, and only after that are translated to matrices. Let us formulate a new property that we call \emph{randomization monotonicity}. First, this property is formulated in terms of normalized class adjacency matrices. Second, it essentially extends the constant baseline from one point to the whole range of the measure: we require some desirable (unbiased) behavior not only for random graphs but also for homophilic/heterophilic structures.

Suppose that for a given graph $G$, we randomly sample some fraction of its edges and redraw them randomly (while keeping the degree distribution, as before). In terms of the normalized class adjacency matrix, we replace $C$ by $(1-\ep)\cdot C + \ep \cdot \rand(C)$ for some $\epsilon>0$. Then, it is natural to require that homophily of the graph gets closer to $R_{base}$ since some of its edges become independent of labels. Formally, we require the following.

\paragraph{Randomization monotonicity}  For any $\ep \in (0,1]$ we have:
\begin{align*}
h(C) &< h((1-\ep) \cdot C + \ep \cdot \rand(C)) \le R_{base} \text{ if } h(C)< R_{base}\,, \\
h(C) &> h((1-\ep) \cdot C + \ep \cdot \rand(C)) \ge R_{base} \text{ if } h(C)> R_{base}\,.
\end{align*}

We believe that the development and analysis of non-contradicting properties for the directed case is an important subject for future studies.

\section{Graph Datasets Description}
\label{sec:datasets_description}

Cora, citeseer, and pubmed \citep{giles1998citeseer, mccallum2000automating, sen2008collective, namata2012query, yang2016revisiting} are three paper citation graph datasets. In cora and citeseer, the labels are paper topics, and in pubmed the labels correspond to the type of diabetes addressed in the paper. Coauthor-cs and coauthor-physics \citep{shchur2018pitfalls} are co-authorship graph datasets. Nodes are authors, an edge corresponds to co-authoring a paper, and labels are fields of study. Amazon-computers and amazon-photo \citep{shchur2018pitfalls} are co-purchasing graph datasets. Nodes are products, an edge means that two products are frequently bought together, and labels are product categories. Lastfm-asia is a social graph of music streaming site LastFM users living in Asia \citep{rozemberczki2020characteristic}. Edges correspond to follower relationships, labels are the user’s nationality. Facebook \citep{rozemberczki2021multi} is a graph dataset, where nodes are official Facebook pages, links correspond to mutual likes, and labels are site categories. Github \citep{rozemberczki2021multi} is a graph dataset, where nodes are GitHub users, edges correspond to follower relationships, and binary label tells that a person is either a web or a machine learning developer. Actor \citep{tang2009social, pei2020geom} is a graph dataset, where nodes are actors, edges correspond to co-occurrence on the same Wikipedia page, and the labels are based on words from an actor’s Wikipedia page. Flickr \citep{zeng2020graphsaint} is a graph dataset of images, where labels are image types. Deezer-europe \citep{rozemberczki2020characteristic} is a graph of users of the music streaming service Deezer, where labels correspond to the user’s gender. Twitch-de and twitch-pt \citep{rozemberczki2021multi} are graph datasets of gamers from the streaming service Twitch, where labels correspond to the use of explicit language by the gamers. Genius \citep{lim2021expertise} is a large-scale heterophilous graph dataset. Roman-empire, amazon-ratings, minesweeper, tolokers, and questions \citep{platonov2023critical} are mid-scale heterophilous datasets. 

We transform all graph datasets to undirected graphs and remove self-loops and multiple edges.

\end{document}